\documentclass[journal]{IEEEtran}
\ifCLASSINFOpdf
\else
\fi
\UseRawInputEncoding
\usepackage{cite}
\usepackage{graphicx}
\usepackage[ruled,vlined]{algorithm2e}
\usepackage{pifont}
\usepackage{booktabs}
\usepackage{amssymb}
\usepackage{amsmath}
\usepackage{color}
\usepackage{multirow}
\usepackage[table,xcdraw]{xcolor}
\usepackage{stfloats}

\begin{document}
%

\title{Graph Representation Learning for Infrared and Visible Image Fusion}

%

\author{Jing~Li, Lu~Bai, Bin~Yang, Chang~Li, Lingfei~Ma, and~Edwin R. Hancock~\IEEEmembership{IEEE~Fellow} 
}

%
%

\markboth{Journal of \LaTeX\ Class Files,~Vol.~x, No.~x, x~x}%
{Shell \MakeLowercase{\textit{et al.}}: Bare Demo of IEEEtran.cls for IEEE Journals}
%



\maketitle

\begin{abstract}
Infrared and visible image fusion aims to extract complementary features to synthesize a single fused image. Many methods employ convolutional neural networks (CNNs) to extract local features due to its translation invariance and locality. However, CNNs fail to consider the image's non-local self-similarity (NLss), though it can expand the receptive field by pooling operations, it still inevitably leads to information loss. In addition, the transformer structure extracts long-range dependence by considering the correlativity among all image patches, leading to information redundancy of such transformer-based methods. However, graph representation is more flexible than grid (CNN) or sequence (transformer structure) representation to address irregular objects, and graph can also construct the relationships among the spatially repeatable details or texture with far-space distance. Therefore, to address the above issues, it is significant to convert images into the graph space and thus adopt graph convolutional networks (GCNs) to extract NLss. This is because the graph can provide a fine structure to aggregate features and propagate information across the nearest vertices without introducing redundant information. Concretely, we implement a cascaded NLss extraction pattern to extract NLss of intra- and inter-modal by exploring interactions of different image pixels in intra- and inter-image positional distance. We commence by preforming GCNs on each intra-modal to aggregate features and propagate information to extract independent intra-modal NLss. Then, GCNs are performed on the concatenate intra-modal NLss features of infrared and visible images, which can explore the cross-domain NLss of inter-modal to reconstruct the fused image. We progressively expand the kennel sizes and dilations to increase the receptive field of GCNs, when we extract intra- and inter-modal NLss. Ablation studies and extensive experiments illustrates the effectiveness and superiority of the proposed method on three datasets.
\end{abstract}

\begin{IEEEkeywords}
 Infrared image, Visible image, Image fusion, Graph Convolutional Networks.
\end{IEEEkeywords}

%
\IEEEpeerreviewmaketitle

\begin{figure}[!t]
\centering
\includegraphics[width=3.5 in]{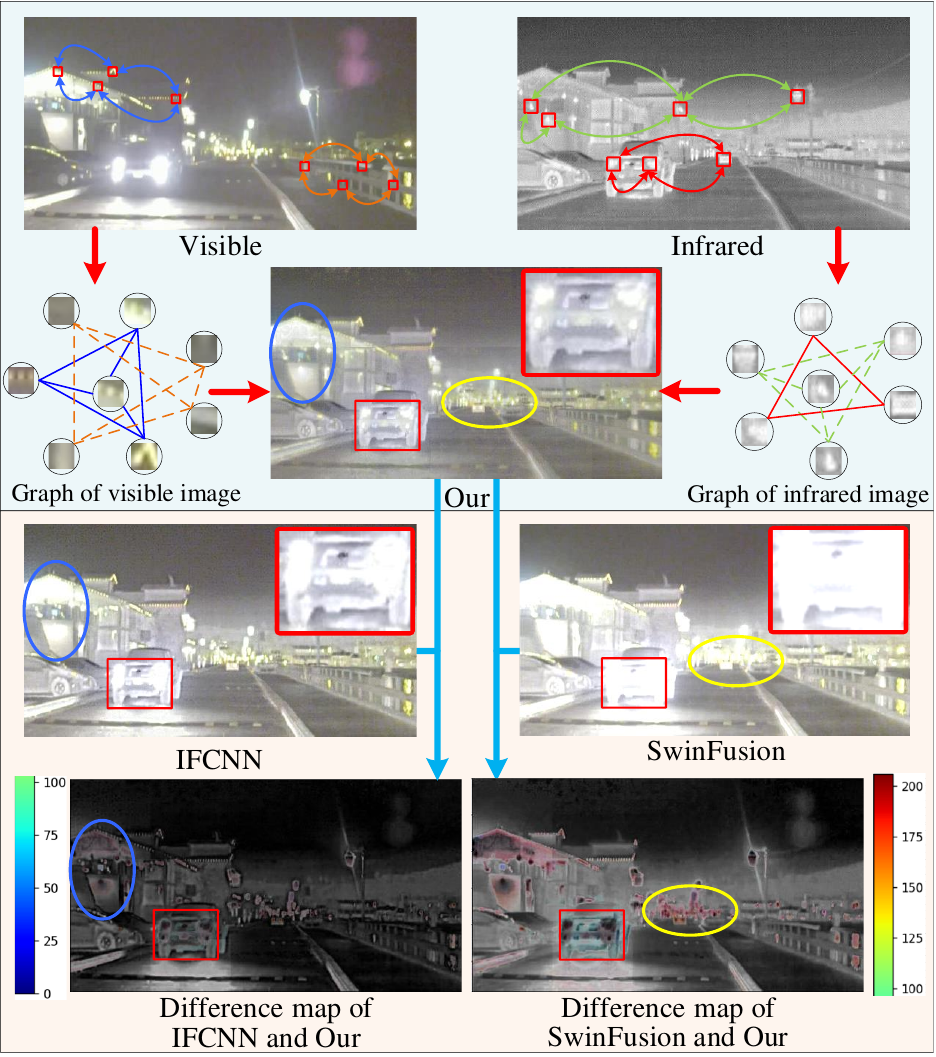}
\caption{Fusion results of our method, CNN-based method (IFCNN \cite{ zhang2020ifcnn }) and transformer-based method (SwinFusion \cite{ma2022swinfusion}). IFCNN mainly focuses on local features, which cannot extract the NLss, such as the spatially repeatable detail or texture of similar objects in far-space distance ($i.e.$, the two endpoint regions of the blue lines in visible image). SwinFusion employs transformer structure, which inevitably leads to redundant information. For example, SwinFusion preserve more illumination information from the lights of cars and street lamps, ($i.e.$, the regions existing in the yellow circles). The difference map also demonstrates that our method can preserve more details than IFCNN and SwinFusion.}
\label{FIG:1}
\end{figure}

\section{Introduction}
\IEEEPARstart{I}{mage} fusion can combine the superiority of multi-source images that captured by
different sensors to produce an informative fused image \cite{karim2022current,zhang2023visible}. Infrared and visible image fusion technology has been widely applied to intelligence acquisition or analysis \cite{ma2019infrared,tang2023datfuse,li2017pixel,ma2019locality,liu2018deep}, because infrared sensors are sensitive to the thermal radiation of the heat source, that can enhance the infrared targets. However, the infrared sensor fails to capture the texture due to the limitation of imaging mechanism. In contrast, the visible sensor can preserve abundant texture by capturing reflected light. Therefore, it is necessary to utilize the advantages of infrared and visible sensors to generate an informative fused image \cite{xu2020u2fusion,tang2022superfusion,zhou2021semantic}. Generally speaking, infrared and visible image fusion can be divided into traditional and deep learning (DL) based methods \cite{li2020attentionfgan,li2019coupled}.

\begin{figure*}[!h]
\centering
\includegraphics[width=6in]{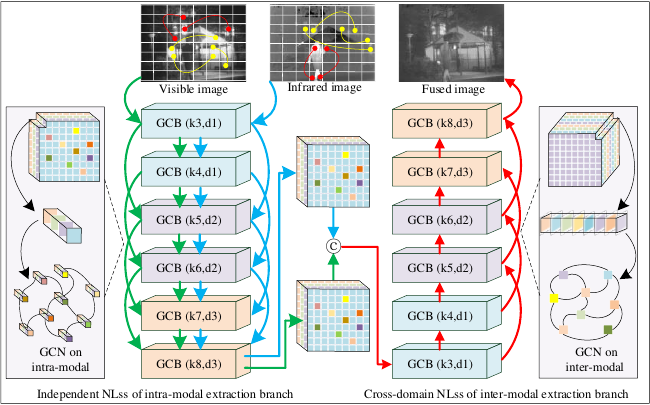}
\caption{The overall framework of the proposed method. We design a cascaded NLss extraction pattern, which first extract independent intra-modal NLss of infrared and visible images by independent NLss of intra-modal extraction branch with two separate data flow, and then cross-domain NLss of inter-modal extraction branch extracts inter-modal NLss by concatenating intra-modal NLss features of infrared and visible images to reconstruct fused image. GCB and \textcircled{c} denote graph convolution block and the concatenate operation, respectively.}
\label{FIG:2}
\end{figure*}

The traditional fusion methods first utilize mathematical transformation to compute features, and then the features are combined by well-designed fusion strategies to generate fused image. Traditional methods mainly include multi-scale transformer methods (MST) \cite{shreyamsha2015image,burt1987laplacian}, sparse representation-based methods \cite{zhang2018sparse}, saliency-based methods \cite{bavirisetti2016two}, subspace-based methods \cite{mitianoudis2007pixel} and other hybrid methods \cite{liu2015general}. Although traditional methods can produce satisfactory results, they still suffer from the problem that the single representation neglects the difference among multi-source images. Moreover, traditional methods also need to manually design complicated fusion strategies.

To overcome the shortcoming of traditional methods, the end-to-end DL models are applied to the fusion task, because DL-based methods can fuse infrared and visible images in an end-to-end way with their powerful nonlinear fitting ability \cite{liu2018deep}. DL methods include CNN-based methods, generative adversarial network (GAN)-based methods, transformer-based methods \cite{xu2022infrared} and the other methods \cite{yue2023dif, ren2021infrared}. CNN-based methods extract infrared and visible image features by designing parallel convolution kernels \cite{tian2021depth}. In addition, the GAN structure is also utilized to image fusion task, which models the distribution of source images by designing an adversarial game. CNN- or GAN-based methods all conduct the sliding window on source images, which can introduce shift-invariance and locality.

Unfortunately, the existing CNN- or GAN-based methods mainly employ convolution operations to extract local features, thus they fail to consider the image's non-local self-similarity. In CNNs, though pooling operations can expand the receptive field to extract global features, it still inevitably leads to information loss. Besides, CNN- or GAN-based methods capture features by designing deeper or complex network without considering long-range dependency. Thus, the under-used latent interactions among image patches, regardless of their intra- and inter-image positional distance, leads to an unsatisfied performance.

Therefore, transformer structure is utilized to address the issues of CNN- or GAN-based methods. Li et al. \cite{li2022cgtf} and Vibashan et al. \cite{vs2021image} combined the transformer with CNNs to extract image's local features and long-range dependencies. In addition, Ma et al. \cite{ma2022swinfusion} and Li et al \cite{wang2022swinfuse} introduced Swin-transformer to infrared and visible image fusion tasks. However, transformer can extract long-range dependencies by self-attention mechanism, yet it considers the correlativity among all the image patches, which leads to transformer-based methods preserve redundant information.

To this end, developing a fusion method to effectively extract non-local self-similarity of images without preserving redundant information still remains a significant challenge. In this paper, we consider that infrared or visible images contain the same or similar irregular objects with spatially repeatable details or texture information. Therefore, it is significant to explore the feasibility of introducing graph representation into infrared and visible fusion task, because graph representation is more flexible than grid (CNN) or sequence (i.e., the transformer structure) representation to deal with the irregular objects, and it can also construct the relationships among the spatially repeatable details or texture information. Concretely, Fig. 1 shows that the different lamps with far-space distance in visible image have spatially repeatable texture information, such as the two endpoint regions of the blue lines in visible image. In addition, the different parts of the car with far-space distance in infrared image have similar thermal radiation details, such as the two endpoint regions of the red lines in infrared image. Similarly, the two endpoint regions of the green or orange lines also have spatially repeatable details. Therefore, when we convert image into graph space, these spatially repeatable regions become the nearest neighbors of themselves such as the graph of infrared or visible image in second row of Fig. 1. Based on this, the graph convolution can aggregate features and propagate information among the nearest vertices of the graph, which can not only extract image's non-local self-similarity, but also flexibly address irregular objects without introducing redundant information.

Fig. 2 shows the overall framework of our model. In our method, we treat the source image as a graph structure rather than a grid or sequence structure, which aims to extract spatial non-local self-similarity relationships and long-range dependency by exploring interactions of different image pixels in intra- and inter-image positional distance. Therefore, we convert images into the graph space, which treats the pixel as the vertices to construct graph structure, then we design a cascaded NLss extraction pattern, which first extracts intra-modal NLss of infrared and visible images by independent NLss of intra-modal extraction branch, and then extracts inter-modal NLss to reconstruct fused image by cross-domain NLss of inter-modal extraction branch. Specifically, in independent NLss of intra-modal extraction branch, we perform GCN on each intra-modal to aggregate features and propagate information across the nearest vertices by two separate data flow, because infrared and visible image have different modals. In cross-domain NLss of inter-modal extraction branch, we perform GCN on the concatenate intra-modal NLss features of infrared and visible images, which aims to increase the interactions of multi-modal features and explore the inter-modal NLss to reconstruct the fused image. Besides, to extract the global features of the intra- and inter-modal, we utilize dynamic graph convolutional networks, which is allowed to dynamically compute the vertex neighbors by the $k$-nearest neighbors (KNN) to construct a new graph structure in each layer. Therefore, we progressively expand the $k$ and dilations of GCN to increase the receptive field, which can extract global features and address the over-smoothing problem in GCN \cite{simonovsky2017dynamic, valsesia2018learning}.

To demonstrate the effectiveness of our fusion method, we also compare our method with a CNN-based method (IFCNN \cite{ zhang2020ifcnn } and a transformer-based method (SwinFusion \cite{ma2022swinfusion}). On the one hand, IFCNN extracts features by CNN, which mainly focuses on local features. Thus, compared with our method, IFCNN cannot preserve spatially repeatable detail or texture information of similar objects in far-space distance ($i.e.$, the two endpoint regions of the green lines in infrared image). For example, IFCNN preserves less details in the blue circles than our method, which is also shown in difference map of IFCNN and our result. On the other hand, SwinFusion employs transformer structure, which inevitably leads to information redundancy in their fused result. For example, SwinFusion preserves too much illumination information from the lights of cars and street lamps, which leads to the detail loss such as the regions existing in the yellow circle and red block. In addition, the difference map of SwinFusion and our result also illustrates that our method can preserve more details than SwinFusion.

Therefore, the key contributions of this work are summarized as follows:
\begin{itemize}
\item To simultaneously extract NLss of images with less redundant information and consider the spatially repeatable detail or texture information of images with similar objects in the far-space distance, we explore the feasibility of introducing the graph representation into the infrared and visible fusion task, which constructs a graph to aggregate features and propagate information among the nearest vertices.

\item To extract the independent modal NLss and cross-domain NLss, we propose a cascaded NLss extraction pattern, which includes two stages, i.e., the intra-modal NLss extraction for feature representation and the cross-domain NLss extraction in inter-modal. The former stage aims to extract the independent intra-modal NLss of each image, and the latter stage can increase the interactions of multi-modal features by extracting the cross-domain NLss to reconstruct fused images.

\item Ablation studies show the effectiveness of our fusion strategies, and qualitative and quantitative experiments demonstrate that our method has better performance than other CNN-, GAN- and transformer-based methods on three datasets.

\end{itemize}

The remainder of this paper is arranged as follows. Related works are presented in Section \uppercase\expandafter{\romannumeral2}. The details of the proposed method are shown in Section \uppercase\expandafter{\romannumeral3}. Section \uppercase\expandafter{\romannumeral4} shows the comparison and generalization experiments. General conclusions are presented in Section \uppercase\expandafter{\romannumeral5}.

\begin{figure}[!h]
\centering
\includegraphics[width=3.5 in]{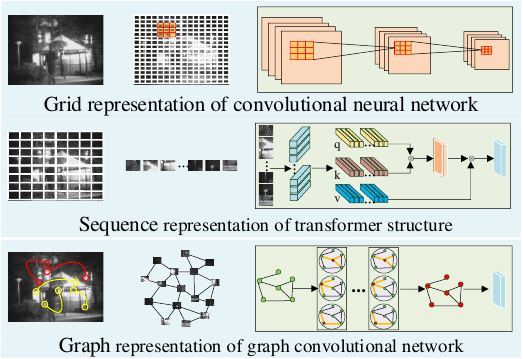}
\caption{The differences of data representation for CNN, transformer and graph convolutional network. We introduce graph representation into infrared and visible fusion task, because graph representation is more flexible than grid (CNN) or sequence (Transformer structure) representation to deal with the irregular objects, and it can also construct the relationships among the spatially repeatable details or texture information with far-space distance.}
\label{FIG:3}
\end{figure}

\section{Related works}
\label{}
This part shows some related works about the CNN-, GAN- and transformer-based methods. In addition, some works of graph convolutional networks for vision tasks are also presented. Fig. 3 shows the differences of data representation for the CNN, the transformer and the graph convolutional network.

\subsection{Image fusion based on CNN and GAN}
CNN fuses images in an end-to-end way to avoid the complicated design of fusion rules that exist in traditional methods. Thus, CNNs were widely applied in infrared and visible image fusion tasks due to  their translation invariance and locality. Zhang et al. \cite{zhang2020ifcnn} utilized CNNs to extract saliency features of source images, and then the features were combined and reconstructed by two convolution layers (IFCNN). Li et al. \cite{li2021rfn} utilized the residual network to address the fusion task with a multi-stage training method (RFN-Nest). Tang et al. \cite{tang2022image} combined the image fusion with the vision task to propose a real-time fusion model via semantic-aware (SeAFusion). Li et al. \cite{li2021different} proposed a meta-learning fusion model to fuse images with different resolutions, and their model can produce fused results with arbitrary resolutions. Jian et al. \cite{jian2021infrared} fused source images by a deep decomposition process and saliency analysis stage to combine local and global saliency regions. In addition, image fusion also can be treated as an unsupervised problem and can be then addressed by GANs. Therefore, Ma et al. \cite{ma2019fusiongan} first utilized GANs to address the image fusion task and proposed FusionGAN, which designs an adversarial game to model the distribution of source images. Then, to better preserve the information of images, many works were proposed by extending GAN to dual discriminators (DDcGAN \cite{ma2020ddcgan} and D2WGAN \cite{li2020infrared}) or multi-classification GAN (GANMcC \cite{ma2020ganmcc}). Moreover, to perceive the discriminative region of source images, an attention mechanism was introduced into the dual discriminators GAN to fuse infrared and visible images (MgANFuse) \cite{li2020multigrained}. However, CNN- or GAN-based methods mainly extracted local features by convolution operations, but ignored the image's non-local self-similarity. Thus, transformer was applied to extract the global features of the image in fusion task.

\subsection{Transformer based fusion methods}
Transformer can capture long-range dependencies via a self-attention mechanism, which has been widely utilized to address sequence data, such as natural language processing \cite{devlin2019bert}. In addition, transformer was also applied in many computer vision tasks by splitting images into patches, and the image patches were embedded to sequence data to extract long-range dependency. Transformer has achieved tremendous success in image classification \cite{zheng2021rethinking}, image segmentation \cite{carion2020end}, and object detection \cite{dosovitskiy2021image}. Therefore, transformer was also applied to infrared and visible image fusion task. Li et al. \cite{li2022cgtf} combined CNN with transformer to extract the local features by CNN and capture long-range dependencies by transformer. Moreover, Wang et al. built a pure transformer network to extract the long-range dependency of images, and they designed a L1-norm based strategy to measure and preserve infrared saliency and visible texture information  \cite{wang2022swinfuse}. Ma et al. \cite{ma2022swinfusion} also proposed a pure transformer based fusion model (SwinFusion), which utilizes the cross-domain global learning to implement intra- and inter-domain fusion based on self-attention and cross-attention, and they introduced Swin transformer to extract long-range dependency of images. However, transformer always considered the correlativity among all the image patches, which leads to the fused result contains redundant information. Therefore, it is necessary to introduce graph representation into infrared and visible fusion task, because graph representation can extract image's non-local self-similarity without introducing redundant information.

\begin{figure}[!t]
\centering
\includegraphics[width=3.5 in]{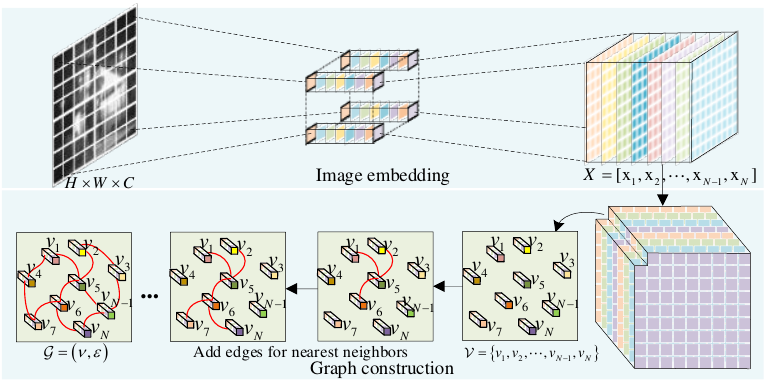}
\caption{Graph representation of the image. In our method, we embed the image patches into feature vectors $X=\left[ {{\text{x}}_{1}},{{\text{x}}_{2}},\cdots ,{{\text{x}}_{N-1}}\text{,}{{\text{x}}_{N}} \right]$, which can be viewed as the vertices $\mathcal{V}=\left\{ {{v}_{1}},{{v}_{2}},\cdots ,{{v}_{N-1}},{{v}_{N}} \right\}$, and then we calculate each vertice's nearest neighbors and add edges to construct the graph representation of the image.}
\label{FIG:4}
\end{figure}

\subsection{Graph convolutional networks for vision tasks}
GCNs have achieved tremendous success in many applications \cite{bai2020learning}. The early GCN architecture was proposed based on the spatial-based GCNs \cite{micheli2009neural}, and then many variants of spatial-based GCN were proposed and applied to different tasks. In addition, Bruna et al. \cite{bruna2013spectral} proposed a spectral-based GCN by utilizing spectral theory, which was also improved and extended by many variants of GCN to deal with graph data. GCN was suitable to tackle graph data, such as natural language processing \cite{bastings2017graph}, and social networks \cite{tang2009relational}. GCNs were also introduced into point cloud data process methods \cite{simonovsky2017dynamic}, because they can analyze the point cloud image by considering their relationships. In computer vision domain, Han et al. \cite{han2022vision} treated images as a graph structure to propose a Vision GNN (VIG), which split images into sub-image patches and regarded them as the nodes of graphs. Thus, VIG can process irregular objects flexibly. Besides, GCN has also been used for hyperspectral image classification \cite{yu2022edge} and instance segmentation \cite{fan2020correlation}. Although several works have applied GCN to deal with multi-feature fusion, such as multi-view learning \cite{chen2023learnable} and hyperspectral image classification \cite{liu2020cnn}, the former mainly fused the features by a fully-connected neural network in the shared latent space of the original multi-view representations, and the latter combined CNN with GCN to extract the small-scale and large-scale spectral-spatial features in pixel- or super-pixel nodes. However, they all failed to consider the NLss of intra- and inter-modals of multi-modal features. Therefore, we introduce the graph representation into the infrared and visible fusion task by adaptively utilizing the neighborhood structure and latent interactions among intra- and inter-image positional distance.

\section{Methodology}
\label{}
This section first introduces the graph representation of the image, followed by some details of the independent NLss of intra-modal / cross-domain NLss of inter-modal extraction branches and loss functions.

\subsection{Graph representation of image}
In our method, we first convert images into the graph space. Fig. 4 shows the construction of graph for an image. Given infrared image and visible image with the size of $H\times W\times C$, $H$, $W$ and $C$ denote the height, weight and channel size of the image, respectively. We first spilt source images to $N$ sub-image patches, and then embed the image patches into feature vectors ${{\text{x}}_{i}}\in {{\mathbb{R}}^{D}}$, where $D$ denotes the dimension and $i=1,2,\cdots ,N$ , we can get $X=\left[ {{\text{x}}_{1}},{{\text{x}}_{2}},\cdots ,{{\text{x}}_{N-1}}\text{,}{{\text{x}}_{N}} \right]$, which can be viewed as the vertices $\mathcal{V}=\left\{ {{v}_{1}},{{v}_{2}},\cdots ,{{v}_{N-1}},{{v}_{N}} \right\}$. In addition, given a vertex ${{\nu }_{i}}$, we calculate its nearest neighbors $\mathcal{N}\left( {{\nu }_{i}} \right)$ and design an edge ${{e}_{ji}}$ from ${{\nu }_{j}}$ to ${{\nu }_{i}}$ for each ${{\nu }_{j}}$ in $\mathcal{N}\left( {{\nu }_{i}} \right)$. To this end, we can transfer the image to the graph structure $\mathcal{G}=\left( \nu ,\varepsilon  \right)$ where $\varepsilon$ denotes the edges of the graph.

\begin{figure}[!t]
\centering
\includegraphics[width=3.5in]{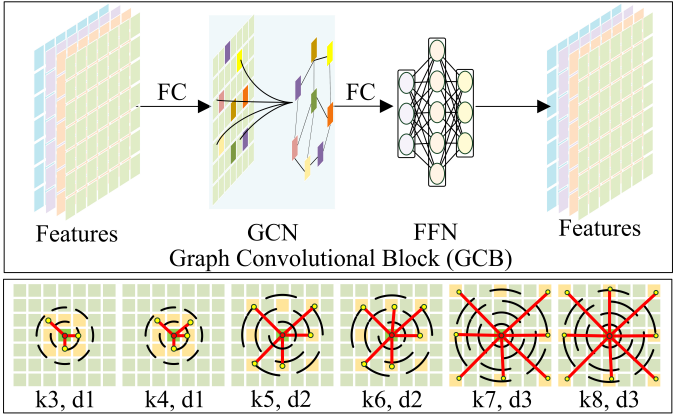}
\caption{The structure of the graph convolution block (GCB) and the details of $k$ and $d$. The $k$ of KNN in graph convolution is progressively expanded from 3 to 8, and the dilation rate of GCNs is progressively expanded from 1 to 3. $k$ and $d$ aim to increase the receptive field.}
\label{FIG:5}
\end{figure}

\subsection{Independent NLss of intra-modal / Cross-domain NLss of inter-modal extraction branches}
In our method, the GCN is conducted on intra-modal and inter-modal to capture spatial non-local self-similarity relationships and long-range dependency. For GCN on the intra-modal, we design an independent NLss of intra-modal extraction branch to extract intra-modal NLss of infrared and visible images, respectively. For GCN on inter-modals, we concatenate the intra-modal NLss features of two source images in channel dimension, and then the features are transferred to a graph structure to reconstruct fused image.

\subsubsection{Graph convolutional operation} in this work, we first embed the image patches into the feature vector $X\in {{\mathbb{R}}^{D}}$, which is utilized to construct a graph $\mathcal{G}=\left( \nu ,\varepsilon  \right)$. To this end, GCN can capture informative features at each vertex by aggregating information from their neighborhoods. The graph convolution operation $\mathsf{\mathcal{F}}$ is defined as follows:
\begin{equation}
\begin{split}
 {{\mathcal{G}}_{l+1}}&=\mathcal{F}\left( {{\mathcal{G}}_{l}},{{\mathcal{W}}_{l}} \right) \\ 
 & =\text{Update}\left( \text{Aggregate}\left( {{\mathcal{G}}_{l}},{{\mathcal{W}}_{agg}} \right),{{\mathcal{W}}_{update}} \right), \\
\end{split}
\end{equation}
where ${{\mathsf{\mathcal{G}}}_{l}}$ and ${{\mathsf{\mathcal{G}}}_{l+1}}$ denote the input and output of the $l$-th layer, respectively. Aggregate is the aggregation function ,which aims to combine information from vertices' neighborhoods. Update denotes the update function, which can calculate the vertex representation by aggregated information. ${{\mathcal{W}}_{agg}}$ and ${{\mathcal{W}}_{update}}$ denote the learnable weights of aggregation and update functions, respectively. In addition, the aggregation and update operations of a vertex exchange information among their neighbor vertices can be formulated as follows:
\begin{equation}
\begin{split}
{{X}_{i+1}}=f\left( {{X}_{i}},h\left( \left\{ {{X}_{j}}|j\in \mathsf{\mathcal{N}}\left( {{X}_{i}} \right) \right\},{{X}_{i}},{{\mathsf{\mathcal{W}}}_{agg}} \right),{{\mathsf{\mathcal{W}}}_{update}} \right),
\end{split}
\end{equation}
where $f\left( \cdot  \right)$ and $h\left( \cdot  \right)$ are the update and aggregation functions, respectively, and the ${{X}_{i}}$ and ${{X}_{i+1}}$ denote the vertex features of $i$-th and ($i+1$)-th layers. $\mathsf{\mathcal{N}}\left( {{X}_{i}} \right)$ denotes the neighbor vertices of a vertex $X$ at $i$-th layer. ${{X}_{j}}$ denotes the parametrized features of its neighbor vertices by ${{\mathsf{\mathcal{W}}}_{agg}}$. Moreover, the aggregation function $h\left( \cdot  \right)$ is formulated as follows:
\begin{equation}
\begin{split}
h\left( \cdot  \right)=\max \left( {{X}_{j}}-{{X}_{i}}|j\in \mathsf{\mathcal{N}}\left( {{X}_{i}} \right) \right).
\end{split}
\end{equation}

Thus, the update function is calculated as $f\left( \cdot  \right)=h\left( \cdot  \right)*{{\mathsf{\mathcal{W}}}_{update}}$. To this end, we adopt the above graph convolution operation in our method, which is denoted as $\text{GraphConv}\left( X \right)$.

\subsubsection{Graph convolution block}
Fig. 5 shows the structure of the graph convolution block (GCB), which aims to dynamically compute the vertex neighbors by the $k$-nearest neighbors (KNNs) and adds the new edge set $\varepsilon _{i}^{dy}$to construct a new graph structure in each layer, which can enlarge the receptive field to extract global features. The $\varepsilon _{i}^{dy}$is defined as follows:
\begin{equation}
\begin{split}
\varepsilon _{i}^{dy}=\left\{ {{e}_{ij}}=\left\{ {{v}_{i}},{{v}_{j}} \right\}|{{v}_{j}}\in \mathsf{\mathcal{N}}\left( {{v}_{i}} \right) \right\},
\end{split}
\end{equation}
where ${{e}_{ij}}$ denotes the edge from ${{\nu }_{i}}$ to ${{\nu }_{j}}$, $\mathcal{N}\left( {{\nu }_{i}} \right)$ is the $k$ nearest neighbors of vertex ${{\nu }_{i}}$. Therefore, the dynamic graph convolution $\text{DynGC}\left( X \right)$ is formulated as follows:
\begin{equation}
\begin{split}
\text{DynGC}\left( X \right)=\text{GraphConv}\left( X,\varepsilon _{i}^{dy} \right),
\end{split}
\end{equation}
where the $X$ denotes the vertex features. In addition, to increase the feature diversity, a fully-connected layer (FC) is utilized before and after graph convolution operations, and the activation function is also applied to avoid model collapse. Moreover, we implement residual learning in the dynamic graph convolution to design a deeper network, which can enable our model to capture more deep features and obtain reliable convergence in training:
\begin{equation}
\begin{split}
X'=\sigma \left( \text{Fc}\left( \text{DynGC}\left( \text{Fc}\left( X \right) \right) \right) \right)+X,
\end{split}
\end{equation}
where $\sigma $ denotes the GeLU activation function \cite{hendrycks2016gaussian} that employed in this work. Besides, to enhance the feature representation of our model, we insert a feed-forward network (FFN), which includes a multi-layer perceptron with two FC layers, and the GCB can be formulated as follows:
\begin{equation}
\begin{split}
\text{GCB}\left( X \right)&=\text{FFN}\left( X' \right) \\ 
 & =\text{FFN}\left( \sigma \left( \text{Fc}\left( \text{DynGC}\left( \text{Fc}\left( X \right) \right) \right) \right)+X \right).
\end{split}
\end{equation}

\subsubsection{Independent NLss of intra-modal extraction branch}
Fig. 2 shows the structure of independent NLss of intra-modal extraction branch, which includes six GCBs with progressively expanded $k$ and dilations $d$ of GCNs to enlarge the receptive field, and $k$ denotes the $k$-nearest neighbors, $d$ represents the dilation coefficient. Fig. 5 also shows the details of the GCN with different $k$ and $d$ values. Besides, we design two separate data flow in this branch to extract intra-modal NLss of infrared and visible images due to their modal discrepancy. Therefore, the intra-modal NLss  $fea_{ir}^{intra}$ of infrared image (ir) and $fea_{vis}^{intra}$ visible image (vis) are formulated as follows:
\begin{equation}
\begin{split}
fea_{ir}^{intra}=\text{GCB}\left( {{X}_{ir}} \right),
\end{split}
\end{equation}
\begin{equation}
\begin{split}
fea_{vis}^{intra}=\text{GCB}\left( {{X}_{vis}} \right),
\end{split}
\end{equation}
where ${{X}_{ir}}$ and ${{X}_{vis}}$ denote the vertex features of infrared and visible images, respectively.

\subsubsection{Cross-domain NLss of inter-modal extraction branch}
the cross-domain NLss of inter-modal extraction branch has the same structure with the independent NLss of intra-modal extraction branch, as shown in Fig. 2. However, this  branch performs GCNs on the concatenate intra-modal NLss features of infrared and visible images to explore the cross-domain NLss. Therefore, we first concatenate the intra-modal NLss of source images in channel wise as follows:
\begin{equation}
\begin{split}
fea^{intra}=\text{Concat}\left( fea_{ir}^{intra} ,fea_{vis}^{intra} \right),
\end{split}
\end{equation}
where $\text{Concat}(\cdot )$ denotes the concatenation operation. Accordingly, we transfer the features ($fea^{intra}$) to a graph $\mathcal{G}$. Furthermore, we utilize GCBs to capture the inter-modal NLss among the inter-image positional distance to reconstruct the fused image, and a convolution operation $\rho \left( \cdot  \right)$ is applied to reduce the data dimension by the following equation:
\begin{equation}
\begin{split}
{{I}_{fused}}=\rho \left( \text{GCB}\left( fea^{intra} \right) \right),
\end{split}
\end{equation}
where ${{I}_{fused}}$ denotes fused image.

\subsection{Loss functions}
To model the data distribution of source images and preserve enough information, we design a loss function to supervise the training process of our method. The total loss $\mathcal{L}$ contains two parts: infrared content loss ${\mathcal{L}_{ir}}$ and visible detail loss ${\mathcal{L}_{vi}}$, which is formulated as follows:
\begin{equation}
\begin{split}
\mathcal{L}={\mathcal{L}_{ir}}+\lambda {\mathcal{L}_{vi}},
\end{split}
\end{equation}
where $\lambda$ aims to balance ${\mathcal{L}_{ir}}$ and ${\mathcal{L}_{vi}}$. The fused image should have a similar intensity to infrared image. Therefore, ${\mathcal{L}_{ir}}$ mainly focuses on the content loss to guide our model to preserve enough infrared intensity information using the following equation:
\begin{equation}
\begin{split}
{\mathcal{L}_{ir}}=\frac{1}{HW}\left( \left\| {{I}_{fused}}-I \right\|_{F}^{2} \right),
\end{split}
\end{equation}
where $I$ and ${{I}_{fused}}$ denote the infrared and fused images, respectively. $H$ and $W$ denote the height and width of an image. $\left\| \cdot  \right\|_{F}^{{}}$ is the matrix Forbenius norm. Besides, to preserve more details from visible image, we not only force the fused image to have similar intensity with visible image, but also calculate gradient discrepancies of the fusion result and visible image to preserve more texture information. Therefore, ${\mathcal{L}_{vi}}$ is defined as follows:
\begin{equation}
\begin{split}
{\mathcal{L}_{vi}}=\frac{1}{HW}\left( \left\| {{I}_{fused}}-V \right\|_{F}^{2}+\left\| \nabla {{I}_{fused}}-\nabla V \right\|_{F}^{2} \right),
\end{split}
\end{equation}
where $V$ denotes visible image, $\nabla $ means the gradient operator.

\begin{figure}[!t]
\centering
\includegraphics[width=3in]{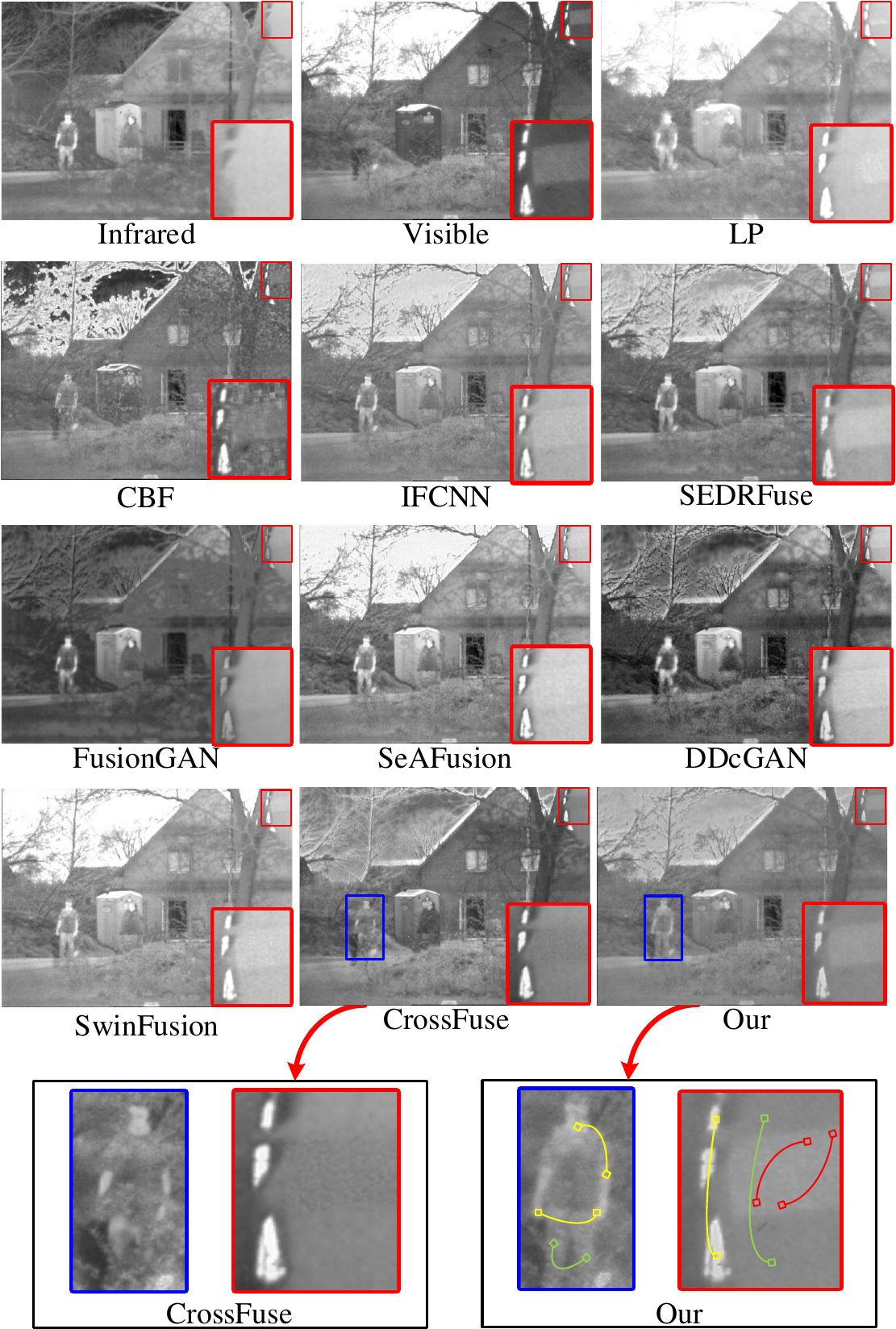}
\caption{Fused example of the proposed method and nine fusion methods on the TNO dataset}
\label{FIG:6}
\end{figure}

\begin{figure}[!t]
\centering
\includegraphics[width=3in]{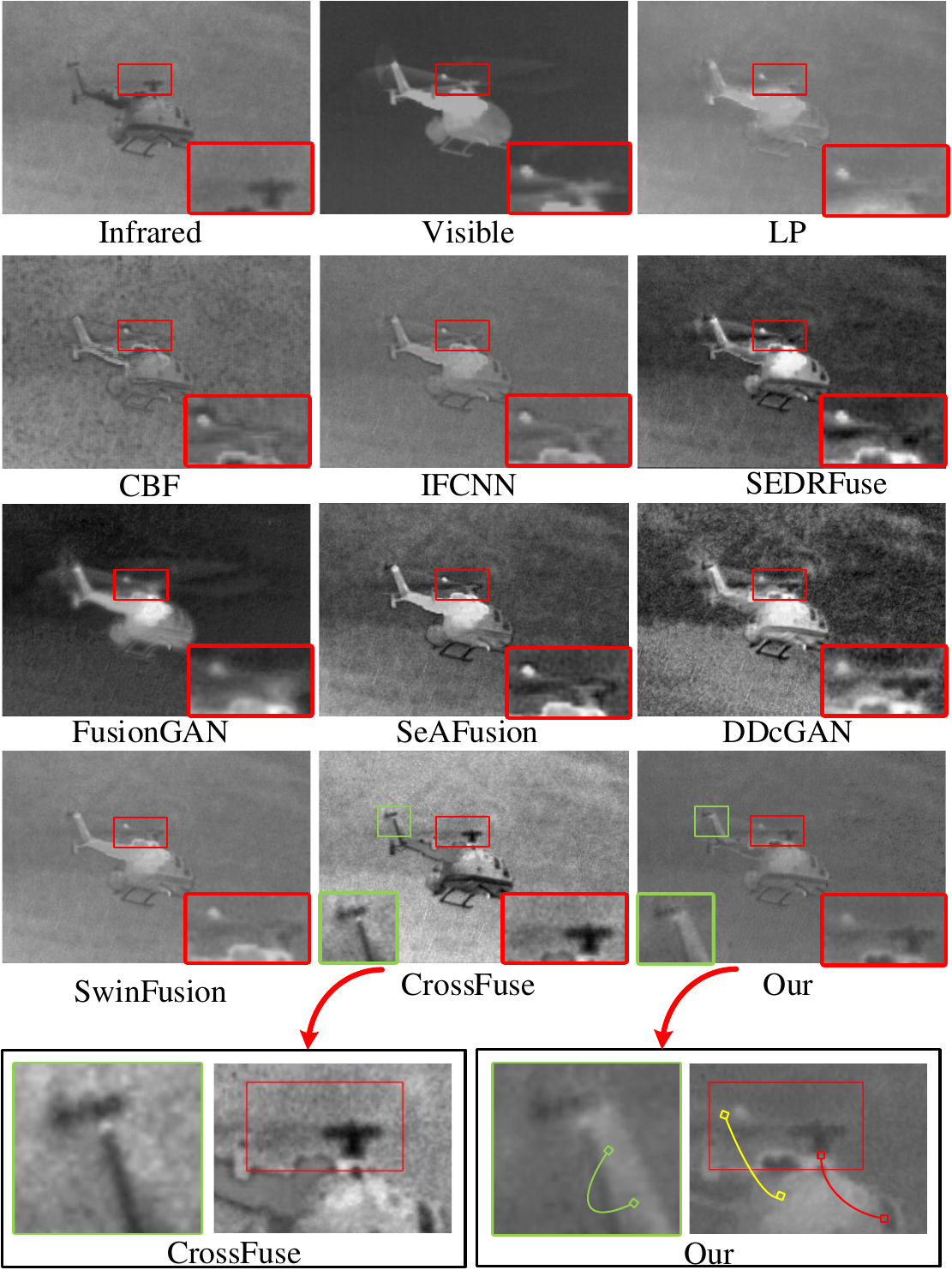}
\caption{Another fused example from the TNO dataset.}
\label{FIG:7}
\end{figure}

\begin{figure}[!t]
\centering
\includegraphics[width=3in]{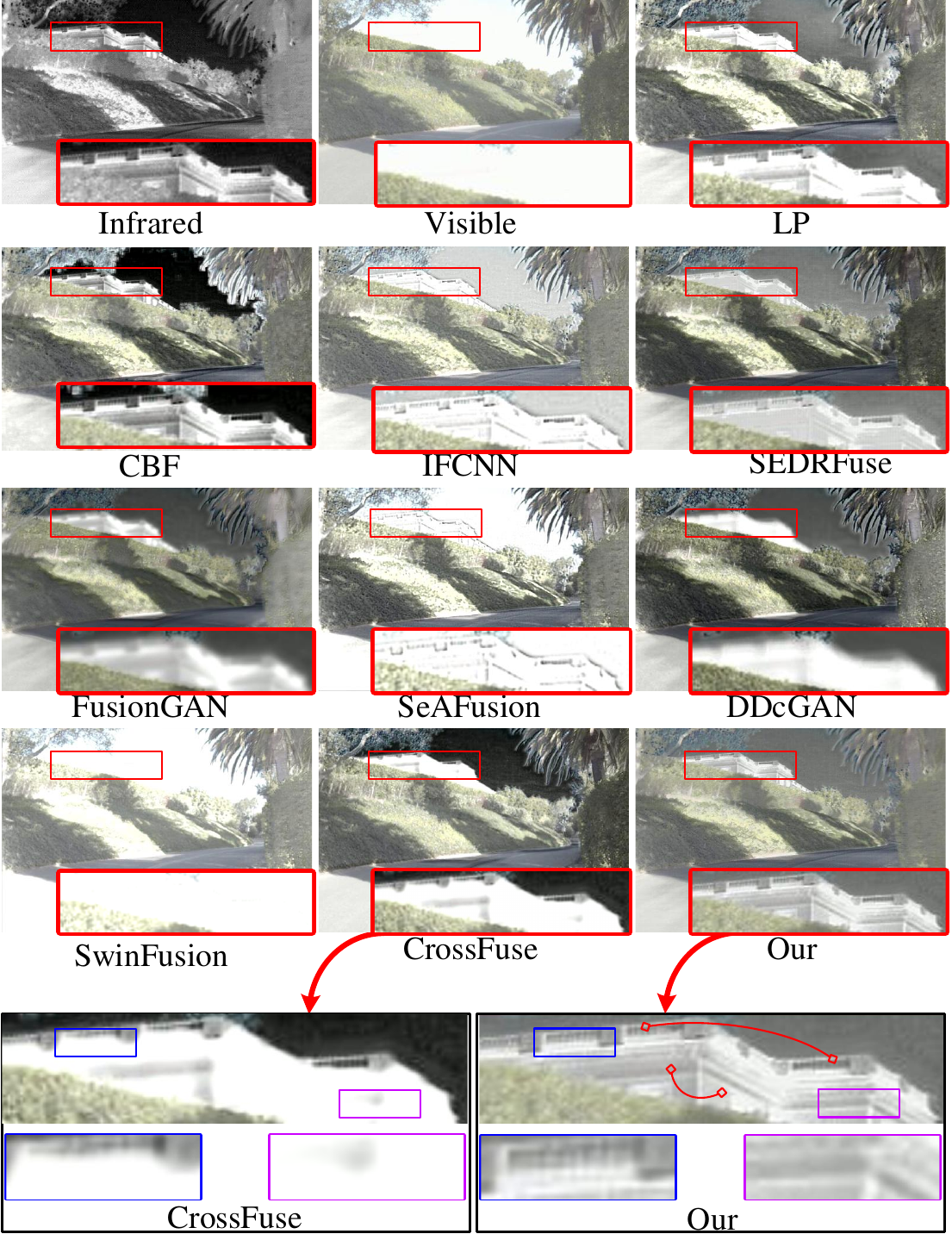}
\caption{Fused example of the proposed method and nine methods on the Roadscene.}
\label{FIG:8}
\end{figure}

\section{Experiments}
In this section, we first show the experimental configurations and some implementation details, and then demonstrate the superiority of the proposed method. We conduct comparative and generalization experiments on TNO  $\footnote{https://figshare.com/articles/TNO\_Image\_Fusion\_Dataset/1008029}$, Roadscene $\footnote{https://github.com/hanna-xu/RoadScene}$ and M3FD $\footnote{https://github.com/dlut-dimt/TarDAL.}$ public datasets in qualitative and quantitative ways. In addition, we also conduct several ablation experiments to demonstrate the effectiveness of cascaded NLss extraction pattern and changeable kennel size and dilation, and we also evaluate our method in the target detection task. In the end, we show the efficiency comparison results of all the compared methods.

\subsection{Experimental configurations}
To comprehensively evaluate the performance of our fusion model, we compare it with nine methods, which include traditional fusion methods (CBF \cite{shreyamsha2015image}, LP \cite{burt1987laplacian}), CNN-based methods (IFCNN \cite{zhang2020ifcnn}, SeAFusion \cite{tang2022image}, SEDRFuse \cite{jian2020sedrfuse}), GAN-based methods (FusionGAN \cite{ma2019fusiongan}, DDcGAN \cite{ma2020ddcgan}, CrossFuse \cite{wang2023cross}) and transformer-based methods (SwinFusion \cite{ma2022swinfusion}). We first conduct the comparison experiments on the TNO dataset, and both infrared and visible images in TNO are gray images. In addition, with the development of sensor technology, the visible sensor does not limit to gray space, which can capture the scene in color space to improve the human visual effect. Therefore, we also select other two public datasets with colorized visible images (Roadscene dataset and M3FD dataset) to illustrate the superiority of our method. For the infrared and color visible image fusion task, we first transform the RGB image to YCbCr, and the Y channel of YCbCr denotes the luminance, which contains the main detail and texture of the scene, Cb and Cr denote the color information. Then, we fuse infrared image and Y channel of visible image to generate a new Y channel ${{Y}^{f}}$. Finally, we combine  ${{Y}^{f}}$ with Cb and Cr channels to produce a color-fused image in RGB space.

In our experiment, qualitative and quantitative analysis are utilized to evaluate the proposed and other methods simultaneously. For qualitative analysis, we evaluate fused results by human visual inspection, which mainly focuses on illuminance, sharpness, contrast and so on. For quantitative analysis, four evaluation metrics are selected in our work, including structural similarity index measure (SSIM) \cite{wang2004image}, peak signal-to-noise ratio (PSNR) \cite{ma2016infrared}, correlation coefficient (CC) \cite{ma2019infrared} and the rate of noise or artifacts added to the fused result in the fusion process (Nabf) \cite{kumar2013multifocus}. Moreover, SSIM aims to measure the structure information among images, PSNR can calculate distortion in the fusion process by computing the ratio of peak value power and noise power in given images, and CC can measure the linear correlation between fused and source images. For SSIM, PSNR and CC, a larger value reflects better performance. However, a smaller value of Nabf denotes better performance.

\subsection{Implementation details}
 In our method, we train the model on the TNO dataset, and we extend it by cropping source images into sub-image with a size of $64\times 64$, and the stride of cropping is set to 20, thus we can collect 22,252 image pairs. The parameters of our method are updated by the Adam optimizer, while the learning rate is initialized to $1\times {{10}^{\text{-}4}}$ and decayed exponentially. In addition, the batch size and epoch of the proposed method are set to 4 and 100, respectively. The $k$ of KNN in graph convolution is progressively expanded from 3 to 8, and the dilation of GCNs is progressively expanded from 1 to 3. Fig. 5 shows the details of different $k$ and $d$ combinations. The hyper-parameters $\lambda$ is set to 1.5 in this study. Besides, we train our model on the NVIDIA P5000 GPU with 32 GB memory.

\subsection{Comparative experiments on TNO}
To comprehensively illustrate the superiorities of our method, we compare it with nine fusion methods on 20 image pairs of TNO in qualitative and quantitative ways. The comparative methods contain traditional methods, CNN-based methods, GAN-based methods and transformer-based methods.

\begin{table*}[]
\centering
\caption{{Quantitative analysis of four metrics on TNO, Roadscene, M3FD datasets, and the pedestrian detection results on M3FD. The \textbf{bold} values indicate the best model performance, and the \textcolor{red}{red} and \textcolor{blue}{blue} values denote the second and third order.}}
\setlength{\tabcolsep}{0.5 mm}
\renewcommand\arraystretch{1.1}{
\begin{tabular}{ccccccccccc}
\hline
Dataset            & \multicolumn{10}{c}{TNO}                                                                                                                                                                                                                                                                                                               \\ \hline
Methods            & CBF                                 & LP                                 & IFCNN                               & FusionGAN                           & DDcGAN       & SEDRFuse                           & SeAFusion                          & SwinFusion                         & CrossFuse   & Our                                 \\ \hline
SSIM ↑             & 0.957±0.322                         & 1.239±0.148                        & {\color[HTML]{00B0F0} 1.293±0.149}  & 1.156±0.146                         & 1.072±0.155  & 1.171±0.174                        & {\color[HTML]{FF0000} 1.306±0.134} & 1.291±0.137                        & 1.191±0.164 & \textbf{1.344±0.14}                 \\
CC    ↑            & 0.389±0.197                         & 0.432±0.193                        & {\color[HTML]{00B0F0} 0.481±0.184}  & 0.418±0.166                         & 0.476±0.156  & {\color[HTML]{FF0000} 0.512±0.169} & 0.477±0.188                        & 0.465±0.194                        & 0.447±0.177 & \textbf{0.513±0.172}                \\
PSNR↑              & {\color[HTML]{00B0F0} 14.638±4.911} & 12.436±3.775                       & {\color[HTML]{FF0000} 14.845±4.218} & 13.347±3.011                        & 12.334±2.228 & 13.606±3.556                       & 13.545±3.719                       & 13.241±3.755                       & 13.659±4.04 & \textbf{15.416±4.364}               \\
Nabf ↓             & 0.132±0.082                         & {\color[HTML]{FF0000} 0.02±0.009}  & 0.062±0.019                         & {\color[HTML]{00B0F0} 0.026±0.03}   & 0.123±0.106  & 0.083±0.081                        & 0.081±0.11                         & 0.051±0.02                         & 0.074±0.086 & \textbf{0.016±0.013}                \\ \hline
Dataset            & \multicolumn{10}{c}{RoadSence}                                                                                                                                                                                                                                                                                                         \\ \hline
SSIM ↑             & 1.256±0.152                         & 1.417±0.137                        & {\color[HTML]{00B0F0} 1.425±0.131}  & 1.196±0.108                         & 1.157±0.115  & 1.198±0.163                        & 1.338±0.109                        & {\color[HTML]{FF0000} 1.444±0.121} & 1.233±0.131 & \textbf{1.456±0.135}                \\
CC    ↑            & 0.595±0.218                         & 0.69±0.166                         & {\color[HTML]{00B0F0} 0.686±0.173}  & 0.63±0.179                          & 0.641±0.14   & 0.685±0.163                        & 0.678±0.18                         & {\color[HTML]{FF0000} 0.691±0.185} & 0.634±0.21  & \textbf{0.692±0.178}                \\
PSNR↑              & {\color[HTML]{00B0F0} 17.179±2.687} & 16.931±2.583                       & {\color[HTML]{FF0000} 17.258±2.486} & 13.081±1.545                        & 14.171±2.019 & 15.354±1.885                       & 15.883±2.038                       & 15.156±2.14                        & 13.96±1.475 & \textbf{17.353±2.522}               \\
Nabf ↓             & 0.037±0.017                         & 0.034±0.007                        & 0.027±0.004                         & 0.044±0.023                         & 0.064±0.024  & {\color[HTML]{FF0000} 0.017±0.01}  & 0.072±0.024                        & {\color[HTML]{00B0F0} 0.02±0.006}  & 0.049±0.015 & \textbf{0.002±0.001}                \\ \hline
Dataset            & \multicolumn{10}{c}{M3FD}                                                                                                                                                                                                                                                                                                              \\ \hline
SSIM ↑             & 0.853±0.233                         & 1.231±0.133                        & {\color[HTML]{00B0F0} 1.327±0.113}  & 1.252±0.13                          & 1.066±0.086  & {\color[HTML]{FF0000} 1.33±0.104}  & 1.306±0.106                        & 1.31±0.11                          & 1.248±0.106 & \textbf{1.339±0.122}                \\
CC    ↑            & 0.342±0.136                         & 0.39±0.131                         & {\color[HTML]{00B0F0} 0.466±0.127}  & 0.365±0.122                         & 0.392±0.114  & \textbf{0.503±0.123}               & 0.434±0.133                        & 0.433±0.139                        & 0.435±0.134 & {\color[HTML]{FF0000} 0.467±0.127}  \\
PSNR↑              & 13.127±2.276                        & 12.265±1.824                       & \textbf{14.371±1.935}               & {\color[HTML]{FF0000} 14.006±1.794} & 11.368±1.105 & 13.598±1.864                       & 13.521±1.976                       & 13.629±2.03                        & 13.16±1.963 & {\color[HTML]{00B0F0} 13.768±1.959} \\
Nabf ↓             & 0.026±0.015                         & {\color[HTML]{FF0000} 0.006±0.002} & 0.028±0.006                         & {\color[HTML]{00B0F0} 0.006±0.005}  & 0.063±0.041  & 0.012±0.01                         & 0.028±0.009                        & 0.026±0.007                        & 0.045±0.015 & \textbf{0.001±0.001}                \\ \hline
                   & \multicolumn{10}{c}{Object detection on M3FD}                                                                                                                                                                                                                                                                                          \\ \hline
AP@0.5             & 0.297                               & 0.191                              & 0.356                               & 0.232                               & 0.072        & 0.353                              & \textbf{0.364}                     & 0.336                              & 0.300       & 0.359                               \\
AP@0.7             & 0.238                               & 0.162                              & 0.291                               & 0.195                               & 0.057        & 0.289                              & \textbf{0.307}                     & 0.281                              & 0.230       & \textbf{0.307}                      \\
AP@0.9             & 0.027                               & 0.025                              & 0.041                               & 0.020                               & 0.007        & 0.043                              & 0.045                              & 0.039                              & 0.029       & \textbf{0.055}                      \\
mAP@{[}0.5:0.95{]} & 0.182                               & 0.124                              & 0.224                               & 0.146                               & 0.043        & 0.225                              & 0.235                              & 0.214                              & 0.182       & \textbf{0.237}                      \\ \hline
\end{tabular}}
\end{table*}

\subsubsection{Qualitative analysis}
Fig. 6 and Fig. 7 show two experimental results of our method and nine fusion methods on the TNO dataset. We select a typical region from each image and zoom in them on the right bottom corner. Compared with nine fusion methods, our method can preserve clearer texture and details. In Fig. 6, all the methods can produce satisfactory results. However, our fused result can capture sharper edges with abundant texture. For example, the regions of the red block of our result in Fig. 6 is more distinct, and it also preserves more texture than other methods, such as the white part in the sign. In Fig. 7, only our result and CrossFuse can clearly present the shape of the propeller of the helicopter in the red blocks. However, our result preserves more meaningful information than CrossFuse, such as the empennage of the helicopter in green blocks.

We also comprehensively compare our method with the transformer-based method (SwinFusion), because transformer can extract the long-range dependency of the image. However, it considers the correlativity among all the image patches by self-attention mechanism, which leads to transformer-based methods inevitably introducing redundant information. For example, the regions that we zoomed in the right bottom of Fig. 6 and Fig. 7 show that our result can preserve clearer texture and capture more details than SwinFusion. Besides, we also zoom in several typical regions of CrossFuse and our results at the bottom of Fig. 6 and Fig. 7, which shows that our result preserves more textures and infrared thermal features than CrossFuse, because we use graph representation to adaptively utilize the neighborhood structure and latent interactions among the regions with spatially repeatable detail or texture information. For example, the red, yellow and green lines that exist in our result denote two endpoint regions of a line that have same or similar objects with spatially repeatable detail or texture information, and our method can aggregate features and propagate information among the nearest vertices rather than all the image patches.

\subsubsection{Quantitative Analysis}
in quantitative experiments, 20 image pairs are selected from TNO to compare various model performance, and we use multiple metrics to evaluate them objectively, which include SSIM, PSNR, CC and Nabf. Table \uppercase\expandafter{\romannumeral1} shows that the proposed method achieves better performance than others on all four metrics. Specifically, our method has the largest SSIM value, which means it can capture more structure features from source images, and the largest PSNR value of our method illustrates it can introduce less noise. Moreover, the largest CC value and the smallest Nabf of our method demonstrate that our method could preserve more information from source images with fewer artifacts.

\begin{figure}[!t]
\centering
\includegraphics[width=3in]{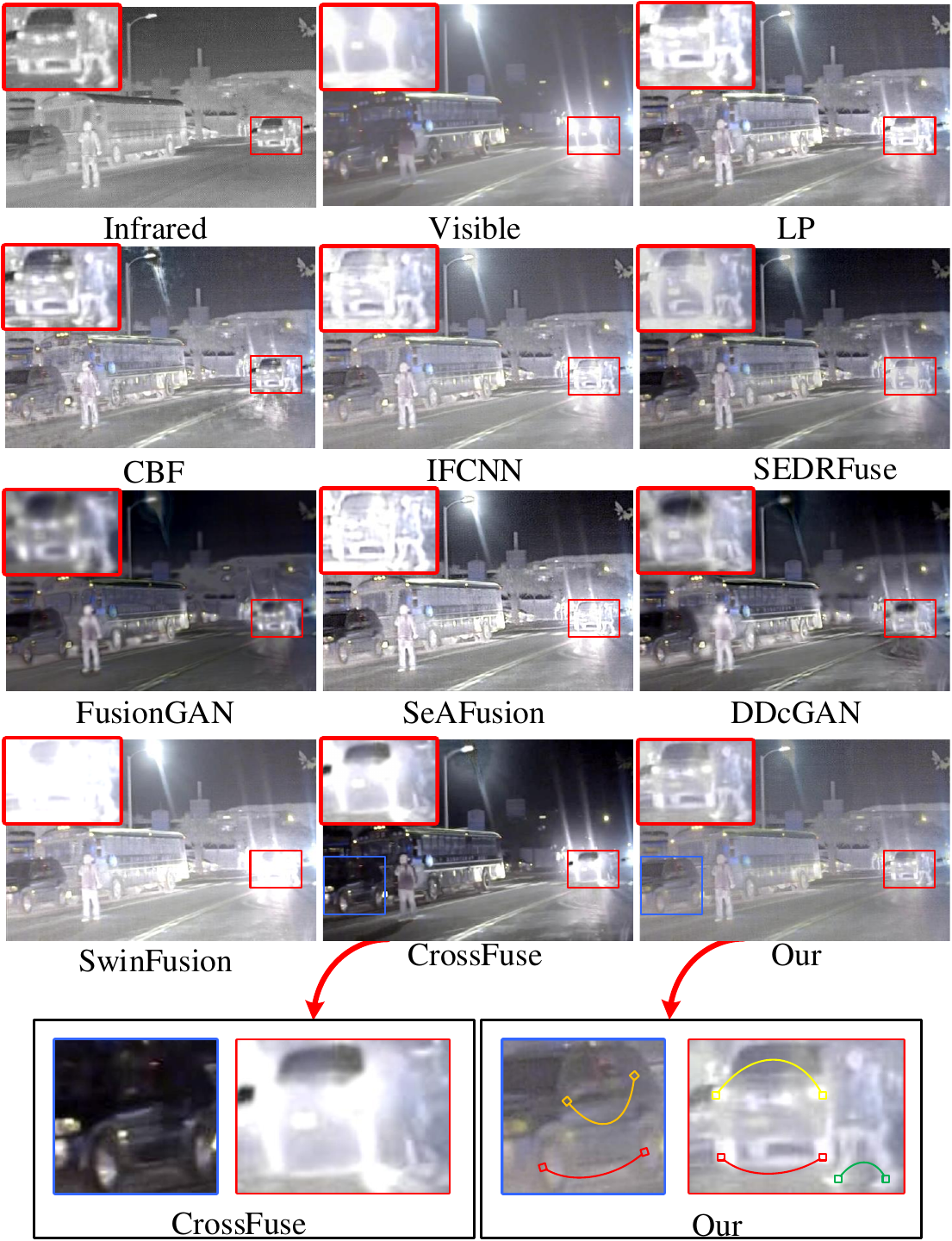}
\caption{Another fused example from the Roadscene.}
\label{FIG:9}
\end{figure}

\subsection{Generalization experiments on Roadscene}
In our experiments, we also analyze the generalization of our method, thus 150 image pairs are selected from Roadscene to test and compare our method with others in qualitative and quantitative ways. In addition, the generalization experiments on Roadscene can also test the performance of our method in RGB space.

\subsubsection{Qualitative analysis}
the visible images in Roadscene are captured in RGB space, thus we show several examples in Fig. 8 and Fig. 9. We also select and zoom in the typical regions in the blue or red blocks to show the differences of fusion methods. Fig. 8 shows that the house existing in the red blocks are missing or blurry in the results of SwinFusion, SeAFusion, CrossFuse, FusionGAN and DDcGAN. However, our result has clear profiles and sharper edges than them, and our result also contains more textures than the other methods. Moreover, the regions existing in the red blocks of Fig. 9 also suffer overexposure issues in SwinFusion, SeAFusion, CrossFuse and IFCNN, which fail to present the details of the car, and our result has clear textures than the others, such as the man in the red block.

SwinFusion and our method all can extract long-range dependency, however our method utilizes graph representation to extract spatial non-local self-similarity relationship without introducing redundant information, Fig. 8 and Fig. 9 show that our results preserve clearer textures than SwinFusion. Besides, we compare our result with CrossFuse, and two typical regions are enlarged at the bottom of Fig. 8 and Fig. 9, which illustrates that our results have clearer texture than CrossFuse, because the results of CrossFuse are overexposure, which leads to the features of the car in zoomed regions are disappeared. The red and yellow lines existing in our results denote two endpoint regions of a line that have same or similar objects with spatially repeatable detail or texture information.

\subsubsection{ Quantitative Analysis}
in generalization experiments, we use four metrics to quantitatively evaluate all the methods on 150 image pairs from the Roadscene. Table \uppercase\expandafter{\romannumeral1} shows that we achieve the best results on all metrics, which illustrates that our method can obtain better results in the quantitative analysis. Therefore, quantitative analysis indicates that our method has better generalization ability on the Roadscene dataset.

\subsection{Generalization experiments on M3FD}
We also conduct a generalization experiment on the M3FD dataset, and 150 image pairs are selected from M3FD to qualitatively and quantitatively evaluate the performance of all the methods.

\begin{figure}[!t]
\centering
\includegraphics[width=3in]{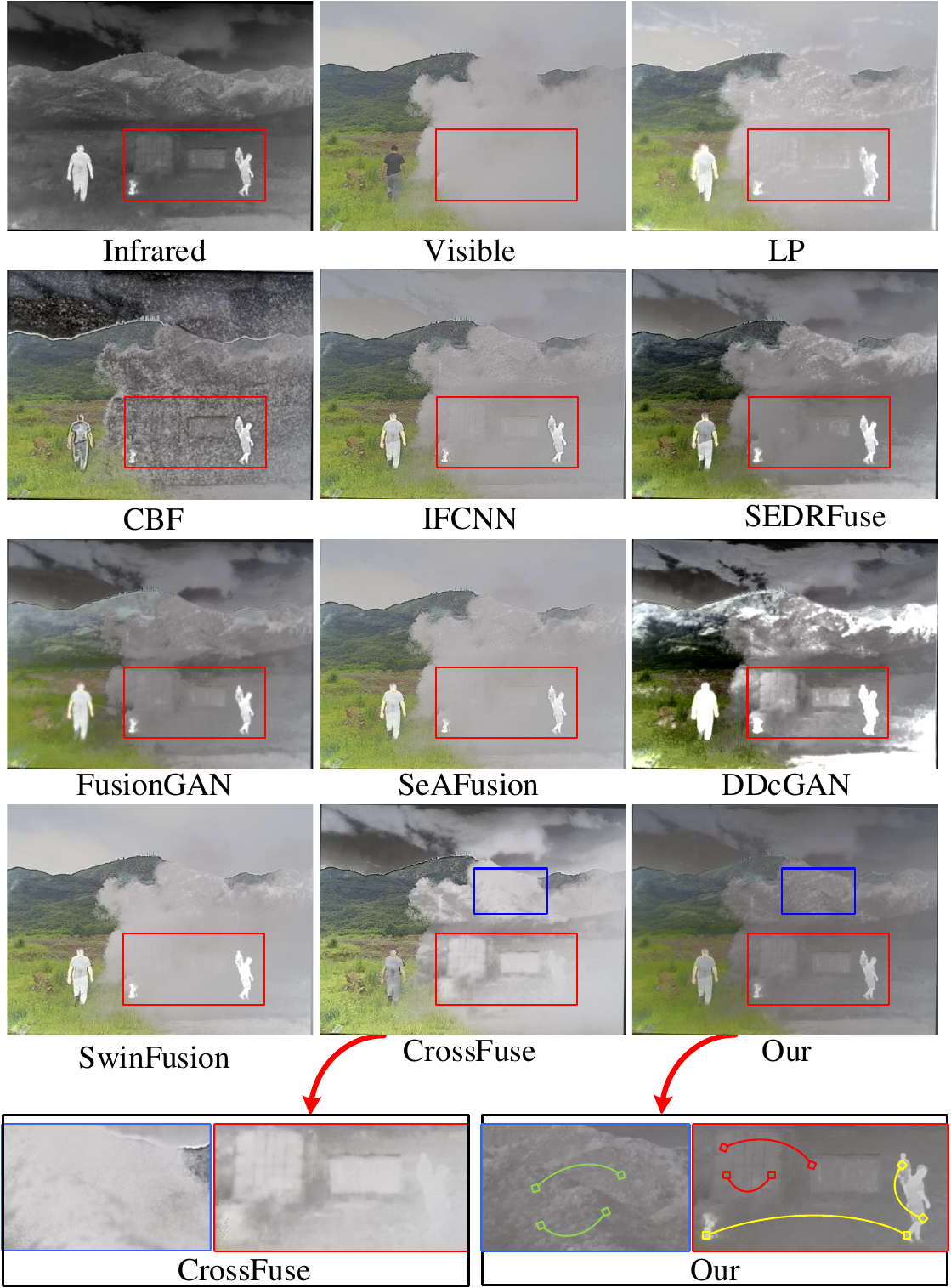}
\caption{Fused example from the M3FD.}
\label{FIG:10}
\end{figure}

\begin{figure}[!t]
\centering
\includegraphics[width=3in]{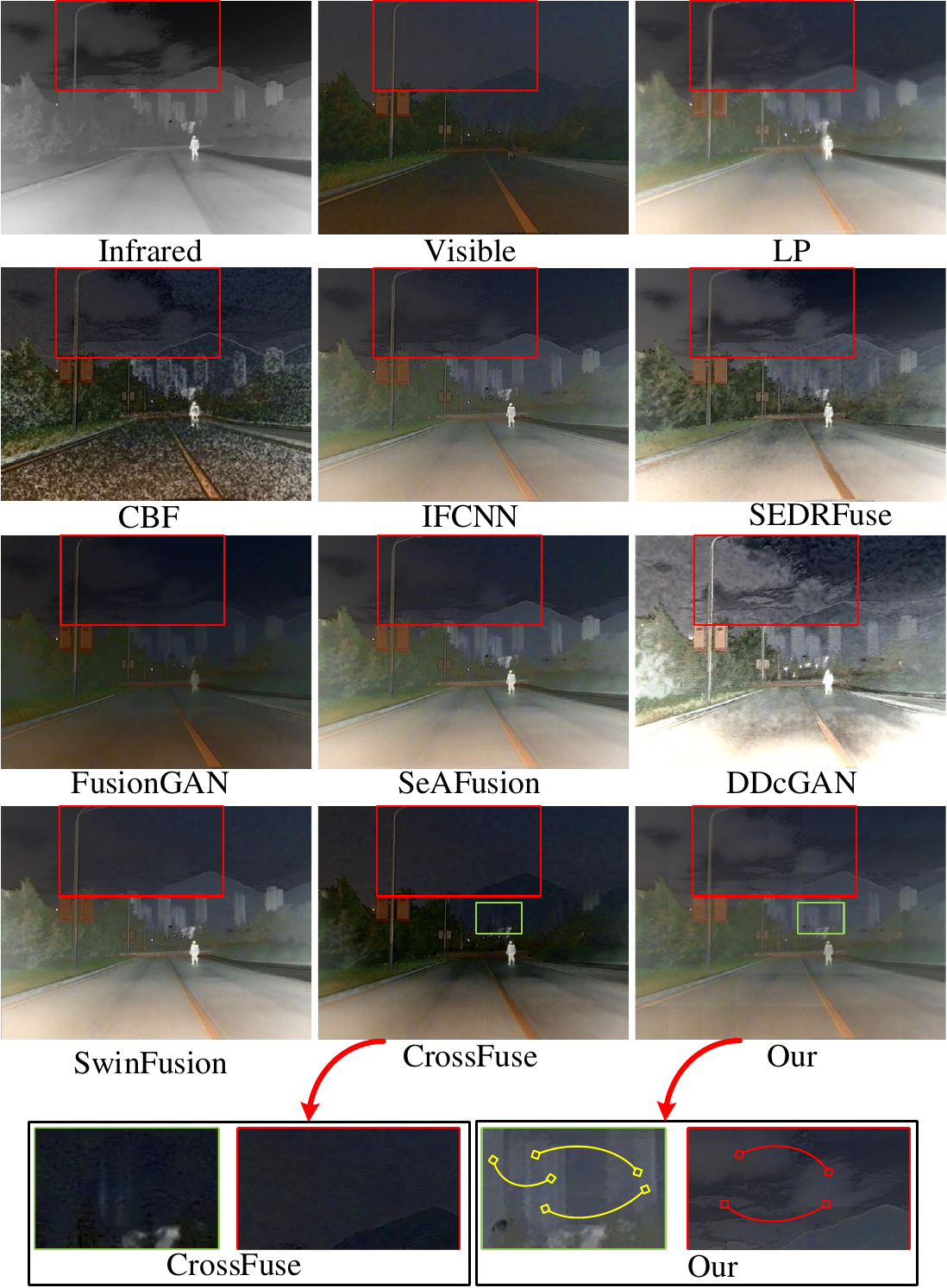}
\caption{Another fused example from M3FD.}
\label{FIG:11}
\end{figure}

\begin{figure}[!t]
\centering
\includegraphics[width=3in]{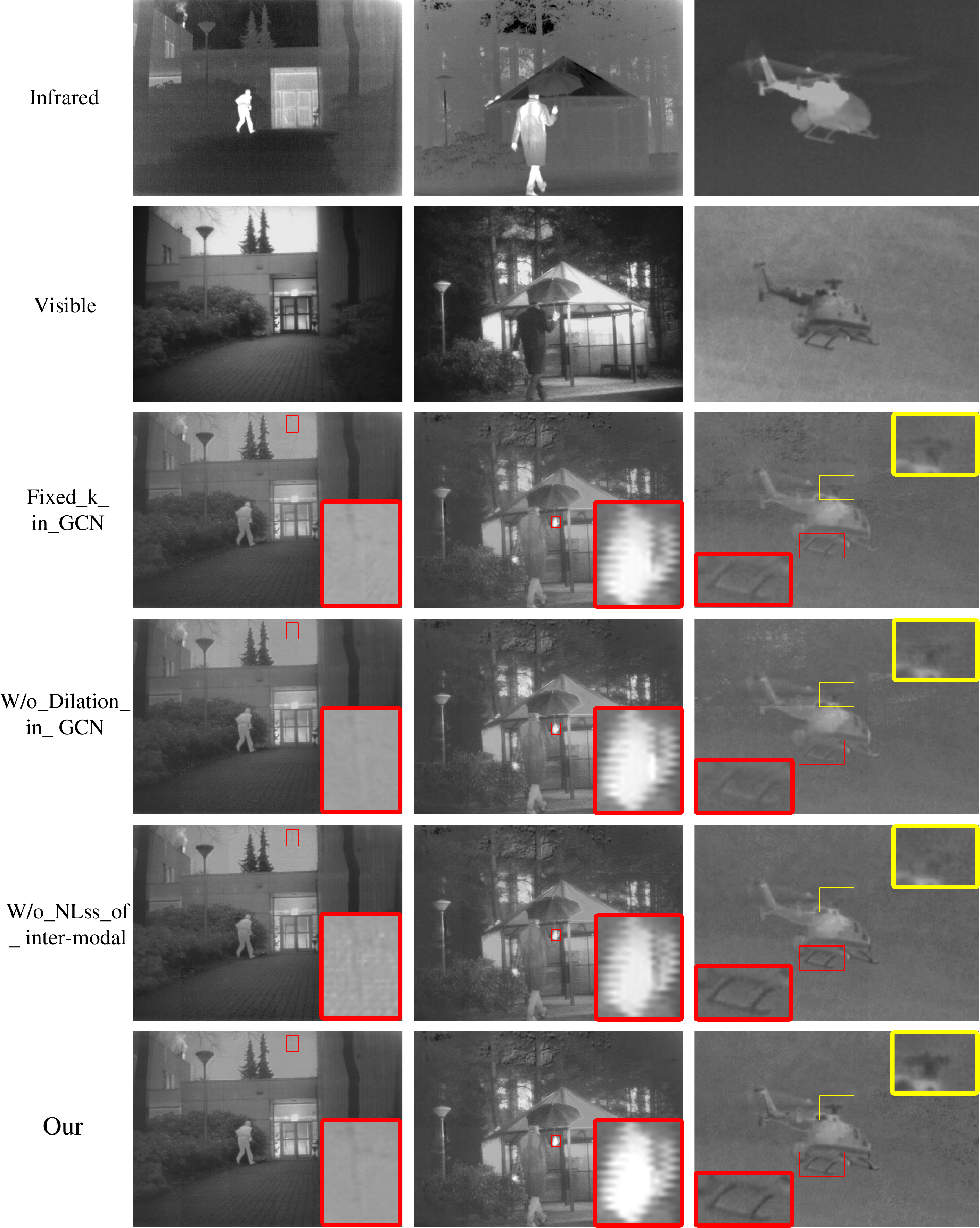}
\caption{Ablation experiments of our method. Fixed\_k\_in\_GCN, W/o\_dilation\_in\_GCN and W/o\_NLss\_of\_inter-modal aim to demonstrate the effectiveness of changeable kennel size, dilation cascaded and NLss extraction pattern.}
\label{FIG:12}
\end{figure}

\subsubsection{Qualitative Analysis}
in Fig. 10, only our method, CrossFuse and DDcGAN can capture the background information that exists in red blocks, such as the buildings. However, our method has more saliency targets than CrossFuse, such as the human in red blocks, and DDcGAN also preserves fewer textures than our method about the human and mountains. Moreover, compared with the other methods of Fig. 10, our result can balance the information of two source images. The qualitative analysis of M3FD illustrates the superiorities of our method.

In addition, we also compare our method with CrossFuse on the M3FD dataset, and several typical regions are selected and presented at the bottom of Fig. 10 and Fig. 11. We can see that our result preserves more details than CrossFuse. For example, the clouds are clear in our result but are disappeared in CrossFuse at the bottom of Fig. 11, and the bottom of Fig. 10 also demonstrates that our method contains more textures and saliency targets than CrossFuse.

\subsubsection{Quantitative Analysis}
four metrics are utilized to quantitatively evaluate all methods on 150 image pairs of M3FD. Table \uppercase\expandafter{\romannumeral1} shows that our method has the largest value on SSIM and also has the minimum performance on Nabf. Besides, it also has acceptable performance on CC (rank second) and PSNR (rank third), which illustrates that we obtain better quantitative results than others in most cases on the M3FD dataset.

\subsection{Ablation analysis}
In this work, we propose a cascaded NLss extraction pattern to extract intra- and inter-modal NLss, and we also progressively expand the kennel size and dilation to increase the receptive field of GCN when we extract intra- and inter-modal NLss. Therefore, to demonstrate the effectiveness of the above operations, we design several ablation experiments.

\begin{figure*}[!h]
\centering
\includegraphics[width=6in]{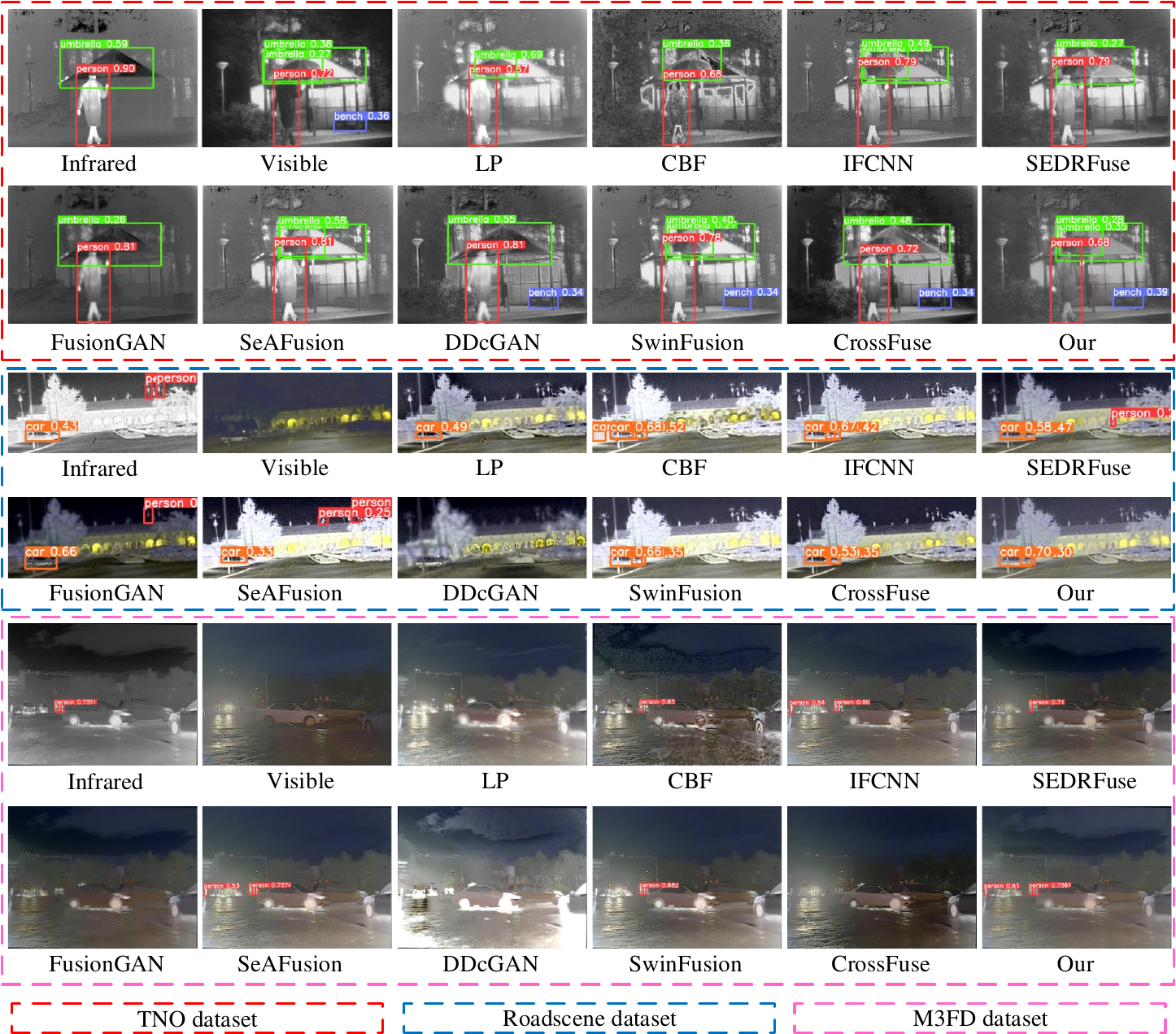}
\caption{The target detection results of the proposed method with 9 methods on three scenes of different datasets}
\label{FIG:13}
\end{figure*}

\subsubsection{Ablation analysis of cascaded NLss extraction pattern}
we implement a cascaded NLss extraction pattern to extract NLss of intra- and inter-modals by exploring interactions of different image pixels in intra- and inter-image positional distance. Specifically, we design an independent NLss of intra-modal branch and cross-domain NLss of inter-modal extraction branch. Therefore, we conduct an ablation experiment to demonstrate the effectiveness of the cascaded NLss extraction pattern by dropping out the cross-domain NLss of inter-modal extraction branch, and the trained model named W/o\_NLss\_of\_inter-modal.

Fig. 12 shows several results of W/o\_NLss\_of\_inter-modal, and our method can preserve more texture information and details than W/o\_NLss\_of\_inter-modal. For example, in the first column, our method preserves clearer texture of the tree that existing in the red block. In the red block of the middle column, our method captures more infrared thermal details of the hand than W/o\_NLss\_of\_inter-modal, and achieves a better visual effect than W/o\_NLss\_of\_inter-modal in the last column, such as the region of the yellow blocks of our method can present the profile and edge of the helicopter. 

In Fig. 12, our method has better performance than W/o\_NLss\_of\_inter-modal. However, W/o\_NLss\_of\_inter-modal also achieves satisfactory fused results, which can also demonstrate the effectiveness of independent NLss of intra-modal branch. Besides, 20 image pairs of TNO are used to compare our method with W/o\_NLss\_of\_inter-modal in a quantitative way, Table \uppercase\expandafter{\romannumeral2} shows that our method has better performance than W/o\_NLss\_of\_inter-modal on all the four metrics, which can demonstrate the effectiveness of the cascaded NLss extraction pattern.

\subsubsection{Ablation analysis of changeable kennel size and dilation}
in our method, to increase the receptive field of GCN when we extract intra- and inter-modal NLss, we progressively expand the kennel size and dilation in different GCB. Thus, we conduct an ablation experiment to demonstrate the effectiveness of changeable kennel size and dilation. Specifically, we train a fusion model by seting a fixed $k$=3, which is named Fixed\_k\_in\_GCN, and also train a fusion model by dropping out dilation operation, which is named W/o\_dilation\_in\_GCN.

Fig. 12 shows the fused results of Fixed\_k\_in\_GCN, W/o\_dilation\_in\_GCN and our method, which demonstrates that our method preserves more texture and infrared thermal objects. For example, only our method can capture the textures existing in the red blocks of the first column, and it also preserves more infrared saliency information than Fixed\_k\_in\_GCN and W/o\_dilation\_in\_GCN in the red block of the second column. In addition, two ablation experiments all fail to preserve the edges and profile of the helicopter, which are captured by our method in the last column.

We also compare our method with Fixed\_k\_in\_GCN and W/o\_dilation\_in\_GCN in a quantitative way. Table \uppercase\expandafter{\romannumeral2} shows the quantitative analysis results, which illustrate that our method has better performance than Fixed\_k\_in\_GCN on all four metrics, and also has better results than W/o\_dilation\_in\_GCN on SSIM, CC and Nabf. The ablation experiments demonstrate the effectiveness of changeable kennel size and dilation.

\subsection{Extended comparison experiments on target detection}
To demonstrate the effectiveness of our fusion model in specific downstream task, we compare our fusion model with other methods in a target detection vision task, which aims to detect targets from digital images, such as pedestrians. Specifically, we design a target detection experiment based on a trained YOLOv5 \cite{redmon2016you} model to evaluate the performance of the proposed and other methods. In the target detection experiment, we take the fused results of each method as the inputs of the trained YOLOv5 model, respectively, and then we compare the performance of target detection to evaluate the effectiveness of fusion models. Fig. 13 shows the target detection results of all the methods, and the three scenes are selected from three datasets, which are located in blocks with different colors. The first two rows present the detection results of the TNO dataset, only our method and SwinFusion can detect all the targets compared to the other methods, yet we achieve better visual effect than SwinFusion. The detect results of Roadscene are presented in the middle two rows, which shows that SEDRFuse, FusionGAN, SeAFusion all detect the person in their result by a mistake. In addition, CBF also causes a false recognition of the car, both LP and DDcGAN fail to detect all the targets. Besides, our method, CrossFuse, SwinFusion and IFCNN all detect the targets, but our result has a higher confidence coefficient and better visual effect than them. The examples of M3FD are shown in the last two rows. Note that LP, FusionGAN, CrossFuse and DDcGAN cannot detect any objects. CBF, ifcnn, SEDRFuse and SwinFusion only detect part of the targets. SeAFusion and our result can detect all the objects, yet our result obtain higher confidence coefficient than SeAFusion. In addition, compared with TNO and Roadscene, M3FD provides the ground turth of pedestrian detection. Therefore, to demonstrate the superiority of our method on the high-level task, we take M3FD to compare the performance of each method by detecting the pedestrian, and the results are shown in Table {\uppercase\expandafter{\romannumeral1}}, which illustrates that our method is superior to other methods in terms of detection AP@0.7, AP@0.9, and mAP@[0.5:0.95]. Therefore, target detection experiments also illustrate the effectiveness and superiority of the proposed method.

\subsection{Efficiency comparison}
In our work, we calculate the average running time of each method on three datasets to compare their efficiency. The traditional methods are implemented by the CPU, and the other methods are conducted by the GPU. Table \uppercase\expandafter{\romannumeral3} shows that CrossFuse consume less time than other methods. In addition, compared with the deep learning based methods, our method consumes more time but less time than CBF, since we need to dynamically compute the vertex neighbors by the k-nearest neighbors (KNNs) and add the new edge set to construct a new graph structure in each layer, which leads to unsatisfied computing efficiency. However, our method can obtain better performance in qualitative and quantitative analysis than the other compared methods. Therefore, considering both computing efficiency and the performance of fused results, our method is still acceptable in most cases.

\begin{table}[]
\centering
\caption{{Quantitative analysis of ablation experiments on four metrics. The bold values indicate the best model performance.}}
\setlength{\tabcolsep}{0.7 mm}
\renewcommand\arraystretch{1.5}{
\begin{tabular}{ccccc}
\hline
                                                                     & SSIM↑               & CC↑                  & PSNR↑                 & Nabf↓                \\ \hline
Fixed\_k\_in\_GCN                                                    & 1.285±0.155         & 0.502±0.172          & 15.342±4.219          & 0.042±0.024          \\
\begin{tabular}[c]{@{}c@{}}W/o\_dilation\_\\ in\_GCN\end{tabular}    & 1.331±0.152         & 0.507±0.175          & \textbf{15.428±4.278} & 0.017±0.009          \\
\begin{tabular}[c]{@{}c@{}}W/o\_NLss\_of\_\\ inter-modal\end{tabular} & 1.264±0.145         & 0.496±0.172          & 15.336±4.285          & 0.036±0.015          \\
Our                                                                  & \textbf{1.344±0.14} & \textbf{0.513±0.172} & 15.416±4.364          & \textbf{0.016±0.013} \\ \hline
\end{tabular}}
\end{table}

\begin{table}[]
\centering
\caption{Mean of running time of proposed method and other methods on three datasets. The bold values indicate the best model performance. (unit: second)}
\setlength{\tabcolsep}{5.5 mm}
\renewcommand\arraystretch{1.1}{
\begin{tabular}{cccc}
\hline
           & TNO           & Roadscene     & M3FD          \\ \hline
CBF        & 10.95         & 5.97          & 57.06         \\
LP         & 0.16          & 0.08          & 1.3           \\
IFCNN      & 0.65          & 0.28          & 1.53          \\
FusionGAN  & 0.41          & 0.49          & 0.31          \\
DDcGAN     & 1.35          & 1.7           & 0.93          \\
CrossFuse  & \textbf{0.04} & 0.03          & \textbf{0.09} \\
SeAFusion  & 0.05          & \textbf{0.02} & 0.19          \\
SwinFusion & 1.62          & 0.78          & 3.97          \\
SEDRFuse   & 2.27          & 1.47          & 6.04          \\
Our        & 3.26          & 2.65          & 7.84          \\ \hline
\end{tabular}}
\end{table}

\section{Conclusion}
In this work, we introduce graph representation to address the issues that the existing CNN- or GAN-based methods capture features by designing deeper or complex networks without considering spatial non-local self-similarity. Therefore, our method transfers images into graph space and proposes a fusion method based on graph convolutional networks, which can adaptively utilize the neighborhood structure and latent interactions among intra- and inter-image positional distance to extract spatial non-local self-similarity of intra- and inter-modal, because the same or similar objects with far-space distance still have spatially repeatable detail or texture information. The qualitative and quantitative analysis of comparative and generalization experiments demonstrate that the proposed method has better performance than the other methods. In future work, we consider to combine graph representation with high-level vision tasks to fuse infrared and visible images, such as the semantic-driven fusion method.

\ifCLASSOPTIONcaptionsoff
  \newpage
\fi



%

\bibliographystyle{IEEEtran} 
\end{document}